\documentclass[11pt, a4paper, logo, copyright]{googledeepmind}

\usepackage[authoryear, sort&compress, round]{natbib}
\usepackage[inkscapeformat=png]{svg}

\usepackage[most, breakable, skins]{tcolorbox}

\tcbuselibrary{skins}
\usepackage{lipsum}
\usepackage{tabularx}
\usepackage{afterpage}
\usepackage{booktabs}
\usepackage{subcaption}
\usepackage{xurl}

\usepackage{makecell}
\usepackage{multirow}
\usepackage{multicol}
\usepackage{array}
\usepackage{epigraph}
\usepackage{float}
\usepackage{listings, listings-rust}
\usepackage{fontawesome5}
\usepackage{amssymb,graphicx}
\usepackage[dvipsnames]{xcolor}
\usepackage{hyperref}
\usepackage[capitalise, noabbrev]{cleveref}
\usepackage{longtable}
\usepackage{graphicx}
\usepackage{ragged2e}
\usepackage{pdflscape}
\usepackage{adjustbox}
\usepackage[section]{placeins}  %
\usepackage{tikz}
\usetikzlibrary{shapes.geometric, arrows, positioning, fit}
\usepackage{tcolorbox} %
\usepackage{listings}
\usepackage{xspace}
\usepackage{caption} %
\usepackage{pifont}

\usepackage{titlesec}
\titlespacing*{\section}{0pt}{*1}{*0.5}
\titlespacing*{\subsection}{0pt}{*0.8}{*0.4}

\newcommand{\ra}[1]{\renewcommand{\arraystretch}{#1}}

\newcommand*{\colorboxed}{}
\def\colorboxed#1#{%
  \colorboxedAux{#1}%
}
\newcommand*{\colorboxedAux}[3]{%
  \begingroup
    \colorlet{cb@saved}{.}%
    \color#1{#2}%
    \boxed{%
      \color{cb@saved}%
      #3%
    }%
  \endgroup
}

\newcommand{\evalname}{\textsl{DeepSearchQA}\xspace}

\title{\evalname: Bridging the Comprehensiveness Gap for Deep Research Agents}

\correspondingauthor{gnikita@google.com, lukhaas@google.com, dipanjand@google.com}

\newcommand{\bSearch}{\ding{171}}
\newcommand{\bGdm}{$\mathbin{\Diamond}$}
\newcommand{\bResearch}{\ding{168}}
\newcommand{\bKaggle}{$\heartsuit$}

\author[*,\bGdm]{Nikita Gupta}
\author[*,\bSearch]{Riju Chatterjee}
\author[*,\bGdm]{Lukas Haas}
\author[*,\bGdm]{Connie Tao}
\author[\bKaggle]{Andrew Wang}
\author[\bResearch]{Chang Liu}
\author[\bGdm]{Hidekazu Oiwa}
\author[\bGdm]{Elena Gribovskaya}
\author[\bGdm]{Jan Ackermann}
\author[\bGdm]{John Blitzer}
\author[\bResearch]{Sasha Goldshtein}
\author[\bGdm]{Dipanjan Das}

\affil[*]{Equal Contribution}
\affil[\bGdm]{Google DeepMind}
\affil[\bSearch]{Google Search}
\affil[\bKaggle]{Kaggle}
\affil[\bResearch]{Google Research}

\begin{abstract}

We introduce \evalname, a 900-prompt benchmark for evaluating agents on difficult multi-step information-seeking tasks across 17 different fields. Unlike traditional benchmarks that target single-answer retrieval or broad-spectrum factuality, DeepSearchQA features a dataset of challenging, hand-crafted tasks designed to evaluate an agent's ability to execute complex search plans to generate exhaustive answer lists. This shift in design explicitly tests three critical, yet under-evaluated capabilities: 1) systematic collation of fragmented information from disparate sources, 2) de-duplication and entity resolution to ensure precision, and 3) the ability to reason about stopping criteria within an open-ended search space. Each task is structured as a causal chain, where discovering information for one step is dependent on the successful completion of the previous one, stressing long-horizon planning and context retention. All tasks are grounded in the open web with objectively verifiable answer sets. Our comprehensive evaluation of state-of-the-art agent architectures reveals significant performance limitations: even the most advanced models struggle to balance high recall with precision. We observe distinct failure modes ranging from premature stopping (under-retrieval) to hedging behaviors, where agents cast an overly wide net of low-confidence answers to artificially boost recall. These findings highlight critical headroom in current agent designs and position DeepSearchQA as an essential diagnostic tool for driving future research toward more robust, deep-research capabilities. DeepSearchQA can be found at \url{https://www.kaggle.com/benchmarks/google/dsqa/leaderboard}.

\end{abstract}

\newread\imgstream
\immediate\openin\imgstream=imagedata.in
\makeatletter
\def\new@kvginclip#1{}
\def\new@kvgintrim#1{}
\let\old@kvginclip\KV@Gin@clip
\let\old@kvgintrim\KV@Gin@trim
\let\oldincludegraphics\includegraphics
\providecommand{\includegraphics}{}
\renewcommand{\includegraphics}[2][]{%
  \immediate\read\imgstream to \src
  \immediate\read\imgstream to \removecrop
  \ifnum\removecrop=1
      \let\KV@Gin@clip\new@kvginclip
      \let\KV@Gin@trim\new@kvgintrim
  \fi
  \oldincludegraphics[#1]{\src}%
  \let\KV@Gin@clip\old@kvginclip
  \let\KV@Gin@trim\old@kvgintrim}
\makeatother

\begin{document}

\maketitle

\section{Introduction}
\label{sec:intro_revised}
\begin{table*}[t]\centering
\caption{Representative examples of user information needs in \evalname. Each row maps a specific domain profile to a complex retrieval task, illustrating the diversity of topics and aggregation requirements.} 
\label{tab:examples}
\scriptsize
\ra{1.3}
\resizebox{0.99\linewidth}{!}{
\begin{tabular}{
p{3cm} p{13.5cm}
}
\toprule
\textbf{Profile} & \textbf{Prompt} \\ \midrule

\textbf{The Epidemiologist} & 
Between 2019 and 2020, which states had a death rate (the number of deaths per 100,000 total population) between 229.8 - < 268.1 that was inflicted by the leading cause of death in the US during 2019 and 2020 per the CDC? \\ \midrule

\textbf{The Video Gamer} & 
I love video games and I'm doing a bit of an analysis into some of the highest selling hardware. Out of the 'PlayStation', 'PlayStation 2', 'PlayStation 3', 'PSP', 'Nintendo DS', 'Nintendo Wii', 'Xbox 360' and 'Game Boy Advance', I want to know which ones sold over 100 million hardware units, include all variations of each that fall under the same family of releases (Use company websites and press releases for this data). Of those which sold more than 100 million units and aren't portable handheld devices, which of these had a 32 bit CPU? And finally, for those devices which satisfy all my constraints, tell me which of their parent company's had assets totalling over 1.76 trillion Yen by the end of the financial year 2010 (March 31st 2010), (use official company financial reports for this data). \\ \midrule

\textbf{The Safety Auditor} & 
According to the NHTSA, which US states had a fatality rate per 100 million vehicle miles traveled below 1 in two or more consecutive years between 2000 and 2005. \\ \midrule

\textbf{The Demographer} & 
Among Canadian provinces and territories that had more than 30 small population centers in 2016, which experienced positive GDP growth and had an Aboriginal identity population exceeding 50,000 in the same year? Use Statistics Canada as the source for all data. \\

\bottomrule
\end{tabular}
}
\end{table*}
The field of artificial intelligence is witnessing a paradigm shift, marked by the rapid transition from static Large Language Models (LLMs) to autonomous web agents designed to interact with dynamic and complex environments to achieve specific goals. This ``agentic revolution'' demands sophisticated capabilities like planning, memory management and tool use. As these systems are increasingly integrated into production-level workflows, tasked with everything from simple data retrieval to complex, multi-step research, the necessity for robust, realistic evaluation methodologies has become paramount. Effective benchmarks are crucial not only for measuring progress but for identifying critical failure modes in these open-ended domains. However, the development of agent capabilities is rapidly outpacing current evaluation methodologies, leading to a significant evaluation bottleneck where existing benchmarks are either saturated or too contrived to represent real-world user needs.

\subsection{The Prevailing Paradigm: Single-Answer Verification}
The previous evaluation methods such as TruthfulQA~\citep{lin2022truthfulqa}, HaluEval~\citep{li2023halueval}, and FELM~\citep{chen2023felm}, or precision benchmarks like SimpleQA~\citep{haas2025simpleqaverifiedreliablefactuality, wei2024measuringshortformfactualitylarge} have successfully established rigorous standards for factuality and single-answer retrieval tasks such as \textit{``What is the capital of France?''}. This single-answer format was a deliberate and highly effective strategy. It enabled low-cost, scalable, and objective automated grading, removing the subjectivity and expense associated with evaluating long-form, generative answers---a massive challenge highlighted by recent work on hallucination and attribution~\citep{rashkin2022measuringattribution,zhao2024wildhallucinationsevaluatinglongformfactuality}. While researchers have developed metrics for long-form consistency---such as FActScore~\citep{min-etal-2023-factscore,Bishop2023LongDocFACTScoreET}, LongFact~\citep{wei2024longformfactuality}, VERISCORE~\citep{song2024veriscore}, and FactAlign~\citep{factalign-longform}---these often require expensive human-in-the-loop verification or complex LLM-based judges~\citep{liu-etal-2024-llms-narcissistic,ramprasad2024automaticfactualitymetricsmeasure}. Similarly, claim verification frameworks like TRUE~\citep{honovich-etal-2022-true}, MiniCheck~\citep{tang-etal-2024-minicheck}, and CoverBench~\citep{jacovi2024coverbenchchallengingbenchmarkcomplex} focus on grounding specific claims rather than evaluating the completeness of a research trajectory.

With the release of reasoning-intensive models~\citep{openai_learning_to_reason,openai2025o3o4, deepseekr1,comanici2025gemini25pushingfrontier,anthropic2025claude4card}, agentic evaluation requires assessing the ability to reason over external information, formulate long-horizon plans, and synthesize findings from unstructured web sources. Previous attempts to address this, such as Humanity's Last Exam (HLE)~\citep{phan2025humanitysexam}, GAIA~\citep{mialon2023gaiabenchmarkgeneralai}, and OpenAI's BrowseComp~\citep{wei2025browsecompsimplechallengingbenchmark}, have pushed the difficulty cieling by introducing expert-level questions that resist simple retrieval. However, even these advanced benchmarks largely retain the single-answer format to ensure objective scoring. Even benchmarks designed for dynamic or search-augmented settings, such as FreshLLMs~\citep{vu2023freshllms} and RealTime QA~\citep{kasai2024realtimeqa}, often rely on atomic answers that do not capture the complexity of broad information gathering.

This single-answer paradigm was a valuable scaffold, isolating core competencies like navigation and search strategy~\citep{miroyan2025searcharenaanalyzingsearchaugmented}. However, it incentivizes precision-first distinct search trajectories ---finding the needle in the haystack---rather than the recall-oriented exhaustive research required in many real-life user information-seeking journeys~\citep{deepresearchbench,openai2025deepresearch}.

\subsection{The Comprehensiveness Gap: Beyond Finding an Answer}
The limitation of the single-answer approach reveals what we term the \textit{Comprehensiveness Gap} in current agent evaluation. Many real-world information-seeking tasks are not satisfied by retrieving a single data point, as seen in traditional datasets like Natural Questions~\citep{kwiatkowski-etal-2019-natural}, TriviaQA~\citep{joshi-etal-2017-triviaqa}, or HotpotQA~\citep{yang2018hotpotqa}. Instead, they require more complex research tasks such as \textit{``List all companies in the semiconductor sector with a P/E ratio under 20 and a presence in Southeast Asia.''} or \textit{``Identify all clinical trials for mRNA vaccines initiated in 2024.''}

This shift from simple retrieval to agentic information retrieval~\citep{zhang2024agentic} necessitates a fundamentally different cognitive profile. It requires three higher-order capabilities that existing benchmarks fail to isolate:
\vspace{-12pt}
\begin{enumerate}
    \item First, the agents must demonstrate \textit{Systematic Collation}, the ability to visit disparate sources --- often hundreds --- and collate a master list where no single source contains the comprehensive information. Recent work on unified retrieval evaluation, such as Fact-Fetch-Reason~\citep{krishna2025factfetch} and MoNaCo~\citep{wolfson2025monaco} begin to address the agentic nature of information retrieval, but lacks the exhaustive scope required here.
    \item Second, agents must perform \textit{Entity Resolution (De-duplication)}. This is the ability to identify when two retrieved entities are identical despite varying surface forms --- a complex task in structured reasoning~\citep{li2024booster} often overlooked in implicit reasoning benchmarks~\citep{geva2021didaristotleuselaptop}. In deep research context, failure to resolve entities leads to inflated lists and degraded precision, a common failure mode we observe in our experiments.
    \item Third, and perhaps most critically, agents must reason about \textit{Stopping Criteria} under epistemic uncertainty. The agent must determine when a search is complete without an explicit termination signal~\citep{yang2024crag}.  This challenges the agent to distinguish between absence of evidence (``I have not found it yet.'') and evidence of absence (``It does not exist.'').
\end{enumerate}
\vspace{-10pt}
\subsection{DeepSearchQA Core Contributions}
\vspace{-5pt}
The primary contribution of \textit{DeepSearchQA} is to shift the evaluation paradigm from precision-based retrieval to exhaustive answer set generation. The benchmark requires agents to conduct deep, autonomous browsing operations to generate a complete, verifiable set of all possible answers for a given query, rather than a single data point.
\begin{figure*}[t]
    \centering
    \includegraphics[width=\textwidth]{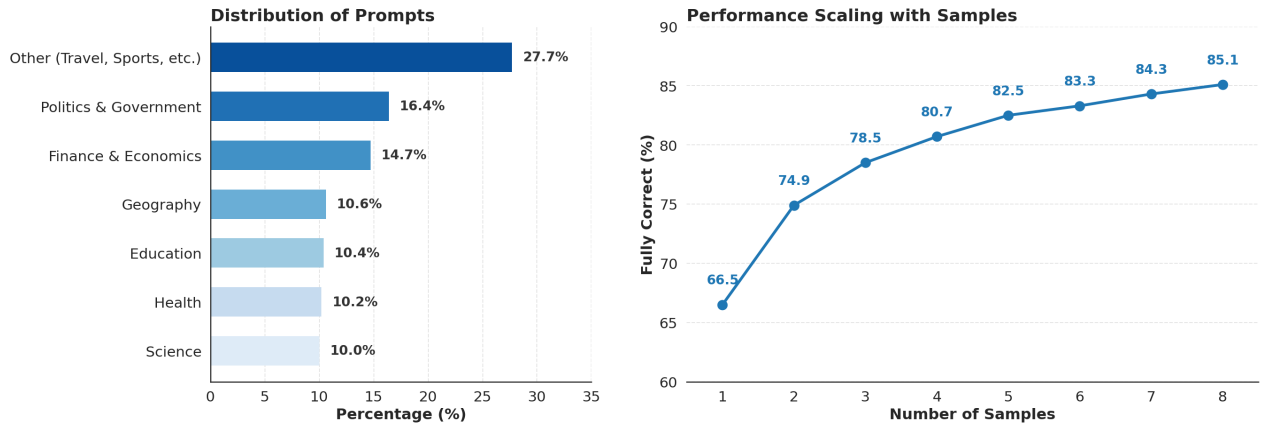}
    \caption{DeepSearchQA benchmark overview. The benchmark features a balanced distribution of prompts across diverse topics (Left), preventing domain overfitting. When evaluated on this diverse set, the Gemini Deep Research agent demonstrates strong performance scaling (Right), with accuracy increasing monotonically as more test-time compute (samples) is applied.}
    \label{fig:distr}
    \vspace{-10pt}
\end{figure*}
In line with the principles of simplicity and verifiability espoused by SimpleQA~\citep{haas2025simpleqaverifiedreliablefactuality, wei2024measuringshortformfactualitylarge}, Facts~\citep{cheng2025thefacts}, and BrowseComp~\citep{wei2025browsecompsimplechallengingbenchmark}, DeepSearchQA employs a strictly outcome-based evaluation methodology. Performance is judged solely on the completeness (recall) and correctness (precision) of the final set submitted by the agent, rather than the search trajectory used to obtain it. This approach encourages architectural diversity while serving as a rigorous functional test. To achieve a high F1 score, an agent is effectively forced to master the trade-off between exploration (casting a wide net) and exploitation (verifying candidates), navigating the open web without a predefined map.

We will maintain a live \evalname leaderboard tracking performance of different models. The leaderboard will remain open to new model submissions. Evaluation will be done by Kaggle.

\begin{table}[t]
    \centering
    \small %
    \renewcommand{\arraystretch}{1.5} %
    
    \begin{tabular}{p{2.5cm}|p{3.0cm}p{4.8cm}p{4.8cm}}
        
        \textbf{Task \newline Complexity} & \textbf{Subtypes}& \textbf{Easy Example} & \textbf{Hard Example} \\ \hline
        
        Structured \newline Retrieval \newline (The "Search") & 
        \raggedright Multi-step search strategy, \newline Obscure info& 
        \textbf{Linear Fact Retrieval} \newline ``What is the capital of France?'' \newline \textit{(Single step, high redundancy)} & 
        \textbf{Dependency Graph Retrieval} \newline ``Which cities in the state where LaMotte established a fort have Amtrak stations?'' \\ 
        
        Context \newline Management \newline(The \newline "Assembly") & 
        \raggedright Full page / doc comprehension & 
        \textbf{Short-Context Extraction} \newline ``Summarize the first paragraph of this news article.'' & 
        \textbf{Massive-Context ``Needle''} \newline ``Based on these 50 uploaded PDF case files, list every instance where a judge cited `Section 404' regarding a liability cap.'' \\ 
        
        Logical \newline Reasoning \newline(The "Thinking") & 
        \raggedright Analytical reasoning, \newline Information synthesis& 
        \textbf{Simple Aggregation} \newline ``List the 5 biggest companies by market cap.'' & 
        \textbf{Inference \& Constraint Solving} \newline ``Identify the US state that has the highest number of Fortune 500 headquarters relative to its population, excluding states that have zero corporate income tax.'' \\ 
        
    \end{tabular}
    \caption{Taxonomy for benchmarking task complexity across domains. The second column describes which subtypes of common cognitive reasoning problems are encompassed by each task. The last two columns compare easy against hard versions of the same examples.}
    \label{tab:task_complexity}
\end{table}
\section{DeepSearchQA: Dataset and Taxonomy}
\label{sec:data}
The \textit{DeepSearchQA} dataset comprises a diverse set of questions curated by expert data annotators from various sources to reflect realistic, high value user queries. These were rigorously filtered to focus on objective, information-seeking tasks where the ground truth is definitive.
\vspace{-5pt}
\paragraph{Dataset Statistics and Domains.}
The benchmark contains 900 prompts paired with ground-truth answer sets. The questions span a broad distribution of domains---including Politics and Government, Finance and Economics, Science, Health, History, Geography, and Media---ensuring that agents are tested on their ability to generalize across varying web structures and content types. \cref{fig:distr} presents the distribution of the domains in the dataset and \cref{tab:examples} shows different examples from our curated dataset. A critical design choice was to ensure that all prompts are time-anchored or reference static data sources (e.g., ``\textit{According to the 2020 Census...}''). This minimizes the drift inherent in live web benchmarks, where the ground truth changes over time.

\begin{table}[t]
    \centering
    \small
    \renewcommand{\arraystretch}{1.5}
    \begin{tabular}{p{7.5cm}|p{7.5cm}}
        \textbf{Full Prompt} & \textbf{Task Complexity} \\ \hline
        
        \textbf{Relocation Planner} \newline
        \scriptsize I'm looking to move house to a new city. Which cities have an average house price of less than £200k according to the 2023 Hometrack 65 city price index? Of those cities, filter down the list to the top 5 English cities with the highest percentage of green space according to Ordnance Survey's 2023 Open Greenspace Data. Filter these cities down to the top 3 by percentage of people economically active and in employment according to the Office for National Statistics annual population survey for the Jan 2022-Dec 2022 period. I also drive and I find the new initiatives make planning car journeys convoluted. Therefore, filter the list down further to cities without a clean air zone introduced before January 2024. & 
        Requires multi-step structured retrieval strategies while maintaining context state across four distinct domain searches to narrow a solution set. \newline \newline
        Filter Housing ($<$£200k) $\rightarrow$ Rank Green Space (Top 5) $\rightarrow$ Rank Employment (Top 3) $\rightarrow$ Exclude Clean Air Zones \\ \hline

        \textbf{Vaccination Study} \newline
        \scriptsize I am researching small countries to use for a case study on vaccinations. Can you give me a complete list of all countries with a population between 1 and 2 million people, a life expectancy of 75 years or more, and a measles immunization rate of at least 85\%? Use World Bank data from 2023, do not include any country that doesn't have immunization data from that year, and just give me a comma separated list of countries. No additional text please. & 
        Requires traversing a dataset to find entities satisfying a complex dependency graph of attributes (Pop $\cap$ LifeExp $\cap$ Vax). \newline \newline
        Filter Population (1-2M) $\rightarrow$ Filter Life Expectancy ($>$75) $\rightarrow$ Filter Immunization ($>$85\%) $\rightarrow$ Extract Intersection \\ \hline

        \textbf{Financial Trend Hunter} \newline
        \scriptsize Refer to macrotrends.net for NVIDIA’s historical annual stock price data and to worldometers.info for United States GDP percentage change. For the years 2020, 2021, 2022, and 2023, tell me which of these saw NVIDIA’s stock price grow by more than 125\% (annual \% change) and the GDP of the United States grow by more than 2.5\%. & 
        Requires synthesizing information from disparate sources and solving mathematical constraints (`Stock > 125\%` AND `GDP > 2.5\%`). \newline \newline
        Retrieve Macrotrends Data $\rightarrow$ Retrieve Worldometer Data $\rightarrow$ Calculate Annual Growth $\rightarrow$ Intersect Thresholds \\ 
    \end{tabular}
    \caption{Representative DeepSearchQA tasks. The analysis illustrates the complexity of queries requiring structured retrieval with constraint verification, context management and cross-domain synthesis.}
    \label{tab:prompt_taxonomy_map}
    \vspace{-10pt}
\end{table}

\paragraph{Answer Types.}
The ground-truth answer space is categorized into two distinct types: 
\begin{itemize}
    \item \textbf{Single Answer:} The answer is a unique entity or value (e.g., a specific date or name). While similar to traditional QA, these require deep research due to the obscurity of the requested data and possible conflicting evidences on the web.
    \item \textbf{Set Answer:} The answer is a collection of items. This includes \textit{Enumeration} (listing all items matching a constraint) and \textit{Composite} responses (answering multiple sub-questions).
\end{itemize}

\paragraph{Quality Verification Protocol.}
To ensure ground truth accuracy, we implemented a standardized three-phase verification protocol. The process included a 1) \textit{Independent Research Phase}, where three reviewers attempted to independently research the solution without access to the ground truth. This was followed by 2) \textit{Verification and Comparison}, where the independent answer was cross-referenced with the original ground-truth answer given by the prompt curator. Any discrepancies triggered a final 3) \textit{Conflict Resolution Phase}, to determine if the original ground-truth answer required updating, the reviewer was incorrect, or the prompt itself was ambiguously flawed. All ambiguous prompts were filtered at this stage.
\vspace{-15pt}
\paragraph{Task Complexity Taxonomy.}
We categorize tasks based on the cognitive and operative demands they place on the agent. This taxonomy helps diagnose specific weaknesses in agent architectures. 
\vspace{-10pt}
\begin{itemize}
    \item \textbf{Structured Retrieval (``The Search''):} This requires devising multi-step search strategies to retrieve obscure, niche information to fulfill the user need.
    \item \textbf{Context Management (``The Assembly''):} This involves ingesting and synthesizing large volumes of information of potentially different formats, where the key challenge is about effectively managing context window limitations.
    \item \textbf{Logical Reasoning (``The Thinker''):} This requires abstract deduction, planning, conflict resolution, and infer latent information from incomplete data.
\end{itemize}
\cref{tab:task_complexity} shows one example for each of the tasks and illustrates the difference between easy examples from previous benchmarks compared to hard examples which we include in our benchmark and \cref{tab:prompt_taxonomy_map} illustrates the required reasoning to solve prompts from our new benchmark.

\section{Evaluation Methodology}
\label{sec:metrics}

The evaluation of DeepSearchQA requires a methodology that can robustly and fairly assess the quality of a set-based answer format, where the order of items is irrelevant and semantic variations in answers are common.

\subsection{Formal Evaluation Metrics}

To formally evaluate performance, we adopt standard metrics from the field of information retrieval as well as introduce several categorical metrics. In the following definitions, let $S_i$ be the set of distinct answers submitted by the agent for a given prompt $i$, and $G_i$ be the ground-truth set of correct answers for that prompt. We utilize a two-tiered evaluation approach: \textit{Continuous Metrics} for granular performance analysis, and \textit{Categorical Classification} for strict success rates.

\paragraph{Continuous Metrics.}
We report standard retrieval metrics to quantify the trade-off between accuracy and exhaustiveness. These metrics are calculated for each individual prompt and then averaged over the entire evaluation set. \textit{F1-Score} ($F_1$) serves as the primary ranking metric, providing a balanced measure of performance by calculating the harmonic mean of Precision ($P$) and Recall ($R$).

For a specific prompt $i$:
    \begin{equation*}
        F_{1,i} = 2 \cdot \frac{P_i \cdot R_i}{P_i + R_i}
    \end{equation*}
    Here, precision ($P_i = \frac{|S_i \cap G_i|}{|S_i|}$) represents the accuracy of the submitted items, while Recall ($R_i = \frac{|S_i \cap G_i|}{|G_i|}$) measures the exhaustiveness of the submission against the ground truth.
    \begin{itemize}
        \item For single-answer tasks: The F1 score is equivalent to exact match accuracy (binary 1.0 or 0.0).
        \item For set-answer tasks: The F1 score ensures performance reflects both completeness and accuracy. This metric strictly penalizes agents that attempt to maximize recall by simply outputting a large number of guesses (hallucinations or ``drift'').
    \end{itemize}

\paragraph{Categorical Classification.}
We also report strict, binary categorizations of each response into one of four disjoint categories based on the set relationship between $S_i$ and $G_i$:

\begin{enumerate}
    \item \textbf{Fully Correct ($S_i = G_i$):} 
    A response is fully correct if and only if the submitted set is semantically identical to the ground-truth set. The agent must identify all correct answers while including zero incorrect answers.
    \begin{itemize}
        \item Single-answer tasks: The agent provides the exact, unique golden entity required.
        \item Set-answer tasks: The agent retrieves the complete, exhaustive list of items ($R_i = 1.0$) without hallucinating any extra items ($P_i = 1.0$).
    \end{itemize}

    \item \textbf{Fully Incorrect ($S_i \cap G_i = \emptyset$):} 
    A response is fully incorrect if the submitted set contains zero correct items. The intersection between the submitted answer and the ground truth is empty.
    \begin{itemize}
        \item Single-answer tasks: The agent provides a wrong answer or claims no answer exists.
        \item Set-answer tasks: The agent fails to retrieve even a single valid item from the ground truth answer set. This suggests a complete failure of the search strategy or a fundamental misunderstanding of the prompt.
    \end{itemize}

    \item \textbf{Partially Correct ($\emptyset \neq (S_i \cap G_i) \subset G_i$):} 
    A response is partially correct, if it contains some but not all correct answers and potentially some not contained by the ground truth. This error mode only exists for set-answer tasks.

    \item \textbf{Correct with Extraneous Answers ($G_i \subset S_i$):} 
    A response falls into this category if the agent successfully identifies all items from the ground-truth answer set ($R_i = 1.0$) but also includes one or more incorrect items ($P_i < 1.0$). This represents the \textit{hedging} failure mode.
    \begin{itemize}
        \item Single-answer tasks: This is treated as a failure. Providing multiple candidates when a unique entity is required implies a failure to disambiguate.
        \begin{itemize}
            \item Example (Conflict Avoidance): \textit{Q: ``Who won the 1994 World Cup?'' (GT: Brazil)} $\rightarrow$ Agent: ``Brazil and Italy.'' The agent lists both the winner and runner-up rather than committing to one.
            \item Example (Popularity Bias): \textit{Q: ``What is the capital of Australia?'' (GT: Canberra)} $\rightarrow$ Agent: ``Canberra or Sydney.'' The agent includes the most famous city alongside the actual capital to mitigate the risk of user error.
        \end{itemize}

        \item Set-answer tasks: The agent identifies all correct items ($R_i = 1.0$) but fails the \textit{stopping criteria}, effectively guessing extra items.
        \begin{itemize}
            \item Example (Classification Drift): \textit{Q: ``List the planets in the Solar System.''} $\rightarrow$ Agent includes Pluto alongside the correct eight.
            \item Example (Association): \textit{Q: ``Which US states border Maine?''} $\rightarrow$ Agent: ``New Hampshire and Vermont.'' The agent hallucinates items that are semantically or geographically close to the target.
        \end{itemize}
    \end{itemize}
\end{enumerate}
\vspace{-10pt}
\subsection{Automated Evaluation Pipeline}

We employ an automated \textit{LLM-as-a-Judge} pipeline to determine the semantic equivalence of extracted answers. For every item $s \in S_i$, the judge determines if $s$ is semantically equivalent to any item $g \in G_i$. For evaluation purposes, we use Gemini 2.5 Flash~\citep{comanici2025gemini25pushingfrontier} in a zero-shot setting using the prompt provided in \cref{app:measurement}.

\section{Results and Analysis}
\label{sec:results}

Table~\ref{eval_results_all_slices} presents the comprehensive evaluation performance on a suite of state-of-the-art \textit{Deep Research} agents and reasoning models. The results highlight the scaling behavior of agentic capability and the persistent difficulty of the \textit{Comprehensiveness Gap}. We report four primary indicators: the percentage of \textit{Fully Correct} responses ($S=G$, where the model retrieved the exact ground-truth set), \textit{Fully Incorrect} responses ($S \cap G = \emptyset$, where no relevant items were found), and the aggregate \textit{F1-Score}, which balances precision and recall. We also report \textit{Correct with Extraneous Answers} where models retrieve the correct answer but fail to filter out non-relevant items.

\vspace{-15pt}
\begin{table}[!tp]
\centering
\caption{Main results on DeepSearchQA, independently evaluated by \href{https://www.kaggle.com/benchmarks/google/dsqa/leaderboard}{Kaggle}. We report the relative number of Fully Correct, Fully Incorrect, and Correct with Extraneous Answers. Additionally, we present the F1-Score achieved by each method. The small number to the right of each metric is the 95 percent confidence interval.} \label{tab:main-results-updated}
\begin{tabular}{rlcccc}\toprule
\# & Model & Fully correct $\uparrow$ & Fully incorrect $\downarrow$ & \multicolumn{1}{c}{\parbox{2.2cm}{\centering Correct with \\ Extraneous \\ Answers}}$\downarrow$ & $F_1$ $\uparrow$ \\\midrule

1 & Gemini Deep Research Agent & $\textbf{66.09}$ {\scriptsize $\pm 3.16$} & $\textbf{9.95}$ {\scriptsize $\pm 2.00$} & $10.30$ {\scriptsize $\pm 2.03$} & $\textbf{81.90}$ \\
2 & GPT-5 Pro High Reasoning & $65.18$ {\scriptsize $\pm 3.11$} & $14.13$ {\scriptsize $\pm 2.28$} & $8.12$ {\scriptsize $\pm 1.79$} & $78.98$ \\
3 & GPT-5 High Reasoning & $59.41$ {\scriptsize $\pm 3.21$} & $19.91$ {\scriptsize $\pm 2.61$} & $6.56$ {\scriptsize $\pm 1.62$} & $73.24$ \\
4 & Gemini 3 Pro Preview & $56.56$ {\scriptsize $\pm 3.24$} & $12.78$ {\scriptsize $\pm 2.18$} & $9.89$ {\scriptsize $\pm 1.95$} & $76.86$ \\
5 & o3 Deep Research & $44.24$ {\scriptsize $\pm 3.27$} & $20.09$ {\scriptsize $\pm 2.64$} & $11.74$ {\scriptsize $\pm 2.12$} & $66.45$ \\
6 & o4 Mini Deep Research & $40.36$ {\scriptsize $\pm 3.21$} & $24.19$ {\scriptsize $\pm 2.80$} & $7.80$ {\scriptsize $\pm 1.76$} & $61.76$ \\
7 & Gemini 2.5 Flash & $25.92$ {\scriptsize $\pm 2.86$} & $45.27$ {\scriptsize $\pm 3.25$} & $5.90$ {\scriptsize $\pm 1.54$} & $42.99$ \\
8 & Claude 4.5 Opus & $24.01$ {\scriptsize $\pm 2.89$} & $50.66$ {\scriptsize $\pm 3.39$} & $4.18$ {\scriptsize $\pm 1.36$} & $40.20$ \\
9 & Claude 4.5 Sonnet & $16.04$ {\scriptsize $\pm 2.40$} & $64.25$ {\scriptsize $\pm 3.13$} & $2.90$ {\scriptsize $\pm 1.10$} & $27.85$ \\
10 & Claude 4.5 Haiku & $12.78$ {\scriptsize $\pm 2.18$} & $71.00$ {\scriptsize $\pm 2.96$} & $1.89$ {\scriptsize $\pm 0.89$} & $22.24$ \\

\bottomrule
\end{tabular}
\label{eval_results_all_slices}
\vspace{-10pt}
\end{table}
\paragraph{State-of-the-Art in Deep Research.} \textit{Gemini Deep Research Agent} and
\textit{GPT-5 Pro High Reasoning}~\citep{openai2025gpt5card}  establish the state-of-the-art on this benchmark, although \textit{Gemini Deep Research Agent} likely does so more efficiently, judging by the significantly lower cost\footnote{\href{https://ai.google.dev/gemini-api/docs/pricing\#pricing-for-agents}{Gemini Deep Research Agent pricing} vs. \href{https://openai.com/api/pricing/}{GPT-5 Pro pricing}.}. The results reveal a clear hierarchy: Deep Research Agents outperform their standalone reasoning model counterparts. This confirms that while reasoning models provide the raw intelligence, the iterative loop of an agent is required to close the \textit{Comprehensiveness Gap.}

Notably, \textit{GPT-5 Pro High Reasoning} and \textit{Gemini Deep Research Agent} exhibit different performance profiles. While both have statistically comparable fully correct rates (66.09\% vs 65.18\%), the \textit{Gemini Deep Research Agent} minimizes catastrophic failures more effectively, achieving the  lowest \textit{Fully Incorrect} rate of 9.95\% vs 14.13\% for GPT-5 Pro high reasoning. 
\cref{tab:failures} illustrates common failure modes exhibited by the top-performing models. Notably, these failures manifest at distinct stages of the information processing pipeline. In the first instance, the \textit{Gemini Deep Research Agent} successfully aggregates the raw data but fails to synthesize the final estimate. In the second, \textit{GPT-5 Pro High Reasoning} encounters an unreadable document and fails to recover, effectively halting the process. Finally, the third example demonstrates a constraint violation, where the model fails to filter data points according to the prompt's requirements, indicating a breakdown in the agent's stopping criteria.

The mid-tier consists of \textit{o3 Deep Research} (Fully Correct: 44.24\%) and \textit{o4 Mini Deep Research} (Fully Correct: 40.36\%). Their Fully Incorrect rates rise to 20.09\% and 24.19\% respectively, showing rapid degradation in the ability to maintain coherent long-horizon search strategies.
\vspace{-10pt}
\paragraph{Scaling Behavior and the Reasoning Threshold.}
The \textit{F1-Score} and \textit{Fully Incorrect} metrics highlight a hard \textit{reasoning threshold} required for autonomous research. While the gap between the top models is narrow, there is a precipitate drop-off for smaller reasoning models. \textit{Gemini 2.5 Flash}, drops to an $F_1$ of 42.99\%, roughly half that of the leader. More critically, the \textit{Fully Incorrect} rate spikes to 45.27\%---nearly five times the failure rate of the SOTA agent. This confirms that \textit{DeepSearchQA} tasks are intractable via simple semantic search; they require structured retrieval and multi-step reasoning. Below a certain parameter or compute threshold, agents suffer from trajectory divergence, pursuing entirely incorrect search paths that yield zero relevant results. This finding suggests that scaling down to a cheaper model for research tasks is not a linear trade-off: it is a step-function drop in utility. On the other side of the spectrum, we can observe that by allocating more test-time compute and sampling $n$ times, we can increase the \textit{Fully Correct} rate from 67.18\% ($n=1$) to 85.71\% ($n=8$). Even sampling only twice yields 74.51\%, and with $n=4$, we can already achieve 81.72\%.
\vspace{-10pt}
\paragraph{Metric Divergence: The Last Mile Problem.}
We observe a distinct divergence between the granular F1-Score and strict Fully Correct ($S=G$) success rates. For instance, while the \textit{Gemini Deep Research Agent} achieves an F1-Score of 81.90\%, its strict success rate is 66.09\%. This roughly 15-point gap represents the \textit{``Last Mile Problem''} in autonomous deep research. For \textit{GPT-5 Pro High Reasoning}, the gap is around 13 points. This is driven by two opposing failure modes: \textit{under-retrieval}, where the agent finds the majority of items but misses the ``long tail'' of obscure entities; and \textit{over-retrieval}, where the agent achieves perfect recall ($R=1.0$) but fails to recognize when the search is complete, hallucinating extra items or drifting into adjacent topics. This gap validates the necessity of strict set-based metrics for agent evaluation because high recall is insufficient if the agent lacks the discernment to filter noise. 
\begin{table}[t]
    \centering
    \footnotesize 
    \renewcommand{\arraystretch}{1.5}
    \begin{tabular}{p{5.2cm}|p{4.8cm}|p{4.8cm}}
        \hline
        \textbf{Full Prompt} & \textbf{Failure Mode} & \textbf{Model Response Snippet} \\ \hline
        
        Which EU country, that does not touch the Mediterranean Sea, had \textbf{the least annual transport} of goods through air, sea, inland waterway, rail, and road by thousand tonnes according to Eurostat in 2022 where every transit method actually had an amount of transported goods. If a country didn't use all 5 methods here, discard that country. 
        & \textbf{Quantitative Estimation Error:} \textit{Gemini Deep Research Agent} correctly identifies all countries that satisfy the conditions in the prompt but fails to rank them correctly. It cannot find concrete information and approximates the quantities which creates the error. 
        & ``Even a conservative \textbf{estimate} places this in the tens of millions ... Total \textbf{Estimated} Tonnage''. \\ \hline

        What are the IATA codes for the US major airports, as classified by the US Department of Transportation, which saw improvements of at least one ranking position of their \textbf{On-Time Arrival Rankings} from the 2023 calendar year to the 2024 calendar year, where the on time percentage in 2024 was at least 77.0\% and no greater than 81.5\%? Use Bureau of Transportation Statistics for data. 
        & \textbf{Tool Call Limitation:} \textit{GPT-5 Pro high reasoning} finds an excel file, which supposedly contains all data. However, it states that it can not open it and stops. \textit{Gemini Deep Research Agent} encounters the data as well but continues to search other sources and answers the question partially correct. 
        & ``this exactly from ... major airport’s ... \textbf{rank and on‑time arrival} percentage back to 2003 ... only as an Excel download, and \textbf{I can’t open Excel files directly in this chat.}'' \\ \hline

        Of all the botanists who have papers in the ``papers of individual botanists'' section of the Sherardian Library \textbf{(MSS. Sherard 26-448)}, which ones never held the position of Sherardian Professor of Botany? Please list their full names in the order that their work appears in the Sherardian Library. 
        & \textbf{Stopping Criterion Failure:} \textit{GPT-5 Pro high reasoning} correctly identifies the list of authors on the described work. However, it then does not correctly filter the list according to the constraints from the prompt. 
        & ``Sources and notes: ... lists the named individuals \textbf{with ...} and presents them in the same \textbf{alphabetical} order ...'' \\ \hline
    \end{tabular}
    \caption{Failure modes on DeepSearchQA. \textit{Gemini Deep Research Agent} and \textit{GPT-5 Pro High Reasoning} have common failure patterns. The first column shows prompts that elicit failures; the second column categorizes the specific reasoning or tool limitation; and the third column provides exact text excerpts from the corresponding model demonstrating the error.}
    \label{tab:failures}
    \vspace{-10pt}
\end{table}

\section{Future Work}
\label{sec:future_work}

\subsection{Limitations}
While \textit{DeepSearchQA} offers a robust framework for evaluating comprehensive retrieval, it relies on specific design choices that entail certain limitations. By employing an exclusively outcome-based evaluation, we effectively treat the agent as a black box. In the absence of trajectory data, it is difficult to distinguish between an agent that reasoned correctly and one that arrived at the correct list through inefficient or accidental means (e.g., lucky guessing). Additionally, the \textit{Static Web Assumption}, while necessary for reproducibility, limits the evaluation of ``breaking news'' retrieval where ground truth is volatile. A task's ground truth may become outdated if source websites are removed or their content is significantly altered. This is a prevalent challenge for all benchmarks operating on the live web, necessitating periodic manual reviews and updates to the dataset.

\subsection{Methodological Extensions}
The DeepSearchQA benchmark provides a foundation that can be extended in several promising directions:
\vspace{-10pt}
\begin{itemize}
    \item \textbf{Incorporating Process-Based Metrics:} 
    Future iterations of the benchmark could categorize agent trajectories (e.g., pages visited, query sequences). While scoring would remain outcome-based, this auxiliary data would provide diagnostic insights into failure modes, helping researchers differentiate between retrieval failures, reasoning errors, and synthesis issues.

    \item \textbf{Dynamic and Time-Sensitive Lists:} 
    To test real-time information retrieval, future versions could introduce ``live'' questions where the ground truth is volatile (e.g., ``List all current members of the UK Parliament's Science and Technology Committee''). This would evaluate an agent's ability to handle temporal constraints and fetch fresh data.

    \item \textbf{Weighted Relevance Scoring:} 
    While the current benchmark treats all ground-truth items with equal weight, future extensions could implement graded relevance. By distinguishing between ``core'' and ``peripheral'' answers, the framework can leverage rank-aware metrics such as normalized Discounted Cumulative Gain~(nCDG)~\citep{jarvelin2002cumulated}.
\end{itemize}

\subsection{Implications for Advancements in Agent Architecture}
By explicitly rewarding exhaustive retrieval, we expect DeepSearchQA to catalyze research for agents that possess a new set of sophisticated skills. High performance on this benchmark will likely require:

\begin{itemize}
    \item \textbf{Systematic Exploration Strategies:} 
    Agents must move beyond opportunistic keyword searching toward methodical exploration strategies (e.g., tree-based or graph-based navigation) to ensure that no relevant sub-pages or sections are overlooked.

    \item \textbf{Advanced Information Synthesis:} 
    As agents gather candidates from heterogeneous sources, they require robust mechanisms to merge information, resolve entity ambiguities, and de-duplicate semantically equivalent answers to form a clean final list.

    \item \textbf{Dynamic Stopping Criteria:} 
    Perhaps the most critical capability is the ability to reason about search completeness. Agents must develop a \textit{stopping criterion} to dynamically determine when the retrieved set is likely exhaustive and further searching would be unproductive---a significant step forward in agentic reasoning.
\end{itemize}

\section{Conclusion}
\label{sec:conclusion}

The rapid advancement of LLM-based agents necessitates a corresponding evolution in the methodologies used to evaluate them. While existing benchmarks like Humanity's Last Exam and BrowseComp have been instrumental in pushing the frontiers of agentic reasoning and information discovery, the \textit{Comprehensiveness Gap} remains a critical blind spot. \textit{DeepSearchQA} addresses this by providing a rigorous, set-based evaluation framework that penalizes both under-retrieval and hedging.

\textit{DeepSearchQA} directly measures the skills of systematic exploration, multi-source synthesis, and search completion---competencies that are vital for many real-world research and analysis tasks. The benchmark's outcome-based evaluation protocol, grounded in the established information retrieval metrics of Precision, Recall, and F1-score, provides a robust and flexible framework for assessing performance without constraining architectural innovation.

Our evaluation reveals that even the most advanced models, such as GPT-5 Pro and Gemini Deep Research Agent, struggle to balance the trade-off between recall and precision. By creating a clear, measurable target for exhaustive retrieval, \textit{DeepSearchQA} aims to catalyze the next phase of agent development---moving from agents that can answer a question to agents that can master a topic and mapping an information landscape. Optimizing for this deeper level of comprehensiveness is the critical next step toward building genuinely capable and reliable autonomous web agents.

\section{Contributions and Acknowledgements}
\label{sec:contributions}

\vspace{0.2cm}

\begin{itemize}
\item \textbf{Experimental design:} Nikita Gupta, Riju Chatterjee and Hidekazu Oiwa created the experimental design behind the benchmark and ran all the reported experiments.\vspace{0.2cm}
\item \textbf{Organization:} Connie Tao, Dipanjan Das, Lukas Haas and John Blitzer managed the overall organization of the effort from start to completion.\vspace{0.2cm}
\item \textbf{Early experimentation:} Emily Ingebricson, 
Chang Liu, Megha Mohabey, Wanzheng Zhu, Oscar Chacaltana, Jan Ackermann, Andrew Wang, Michael Collins and Sasha Goldshtein contributed to ideas, conducted data collection and ran experiments.\vspace{0.2cm}
\item \textbf{Sponsors:} Srinivasan (Cheenu) Venkatachary, Koray Kavukcuoglu, Slav Petrov, Ya Xu, and Yossi Matias sponsored the effort and provided technical guidance.
\end{itemize}

All authors wrote parts of the report.\vspace{0.2cm}

We would also like to thank:
\begin{itemize}
    \item \textbf{Gemini team} for the support and model access.
    \item \textbf{Kaggle team} for their expertise, releasing the leaderboard, and running final evaluations.
    \item \textbf{Expert data annotators} who helped to collect examples in the paper.
    \item Our reviewers \textbf{Kristina Toutanova}, \textbf{Jayant Madhavan}, \textbf{Chris Alberti} and \textbf{Phoebe Kirk} for valuable feedback.
\end{itemize}

\bibliography{main}

\clearpage

\appendix

\section{Judge Prompt Templates}
\label{app:measurement}

\noindent\textbf{Grader Prompt}
\begin{lstlisting}
Your task is to evaluate whether a given "AI Response" for a specific "User Prompt" arrived at the correct answer.

**Answer Correctness Task**

*   **Purpose:** Assess whether the AI response provides the correct answer(s) based on the provided "Correct Answer" and "Prompt Type".
*   **Process:**
    *   Identify the "Prompt Type": "<prompt_type>".
    *   Refer to the "Correct Answer": "<answer>".
    *   Based on the "Prompt Type", determine if the "AI Response" contains the expected answer(s).
        *   **'Single Answer'**: Check if the response provides the answer that addresses the user's question. It does not have to match the exact wording of the provided answer.
        *   **'Set Answer'**: Check if the response includes *each* item from the provided ground truth answers. The order might not matter unless specified otherwise. The response might include more answers than the list. Determine the correctness *only* based on the list first and then check if the response includes answers not in the list.
    *   **Explanation:** Provide a brief explanation justifying your assessment of answer correctness, referencing specific parts of the AI response and the correct answer.
    *   **Correctness Details:** Provide a dictionary, one key for each expected answer part, and value is a boolean indicating whether each expected answer part was found.
        *   For 'Set Answer', this will be a list of attributes, one for each item/part in the "Correct Answer". Each key will be a string indicating the expected answer part, and the value will be a boolean indicating whether that part was found in the response.
    *   **Excessive Answers:** Provide a list of strings, each indicating an excessive answer part. If the response provides answers that are **not** in the "Correct Answer" list, add these answers as excessive answers. Return an empty list when there's no excessive answers in the response.


**Output Format:**

Your evaluation *must* be structured as a nested JSON dictionary with the following top-level keys: `"Answer Correctness"`. Please return NULL if any of "Prompt", "AI Response" or "Correct Answer" is empty.
The value for `"Answer Correctness"` should be a dictionary containing `"Explanation"` (a string), `"Correctness Details"` (a dictionary where each key is the expected correct answer, and the value is a boolean indicating whether the response contains the correct answer), and `"Excessive Answers"` (a list of strings indicating the excessive answers).

Make sure you return a valid JSON string. Pay special attention to quotes, commas and special characters in the JSON string. Make sure to escape all special characters and quotes in the JSON string.
\end{lstlisting}

\newpage
\noindent\textbf{Grader Partial Output Example}
\begin{lstlisting}
**Example (Partial):**

"```json
{{
  "Answer Correctness": {{
    "Explanation": "The response correctly identified Belgium and France but also includes an excessive answer, Italy.",
    "Correctness Details": {{
      "Belgium": true,
      "France": true,
    }},
    "Excessive Answers": [ "Italy" ]
  }}
}}
```"

**Now, proceed with the evaluation using the provided User Prompt, AI Response, and Correct Answer.**

User Prompt (Wrapped in <prompt> and </prompt>):
<prompt>
{prompt}
</prompt>
--------------------
**  Correct Answer (Wrapped in <answer> and </answer>):
Prompt Type: {prompt_type}
<answer>
{answer}
</answer>
--------------------
AI assistant response (Wrapped in <response> and </response>):
<response>
{response}
</response>

--------------------
Rating:'''
\end{lstlisting}

\immediate\closein\imgstream

\end{document}